\documentclass{llncs}

\usepackage{graphicx}
\usepackage[utf8]{inputenc}

\begin{document}
%
%
\pagestyle{headings}  
\mainmatter              
\title{Emerging Language Spaces Learned From Massively Multilingual Corpora}
%
\titlerunning{Emerging Language Spaces}  
%
\author{Jörg Tiedemann}
\authorrunning{Jörg Tiedemann} 
%
\tocauthor{Jörg Tiedemann}
\institute{University of Helsinki\\
\email{jorg.tiedemann AT helsinki.fi}}

\maketitle              

\begin{abstract}
Translations capture important information about languages that can be used as implicit supervision in learning linguistic properties and semantic representations. In an information-centric view, translated texts may be considered as semantic mirrors of the original text and the significant variations that we can observe across 
various languages can be used to disambiguate a given expression using the linguistic signal that is grounded in translation. Parallel corpora consisting of massive amounts of human translations with a large linguistic variation can be applied to increase abstractions and we propose the use of highly multilingual machine translation models to find language-independent meaning representations. Our initial experiments show that neural machine translation models can indeed learn in such a setup and we can show that the learning algorithm picks up information about the relation between languages in order to optimize transfer leaning with shared parameters. The model creates a continuous language space that represents relationships in terms of geometric distances, which we can visualize to illustrate how languages cluster according to language families and groups. Does this open the door for new ideas of data-driven language typology with promising models and techniques in empirical cross-linguistic research?
\end{abstract}

\section{Introduction and Motivation}

Our primary goal is to learn meaning representations of sentences and sentence fragments by looking at the distributional information that is available in parallel corpora of human translations. The basic idea is to use translations into other languages as ``semantic mirrors'' of the original text, assuming that they represent the same meaning but with different symbols, wordings and linguistic structures. For this we discard any meaning diversions that may happen in translation due to target audience adaptation or other processes that may influence the semantics of the translated texts. We also assume that the material can be divided into meaningful and self-contained units, Bible verses in our case, and focus on the global data-driven model that hopefully can cope with instances that violate our assumptions.

Our model is based on the intuition that the huge amount of variation and the cross-lingual differences in language ambiguity make it possible to learn semantic distinctions purely from data. The translations are, thus, used as a naturally occurring signal (or {\em cross-lingual grounding}) that can be applied as a form of implicit supervision for the learning procedure, mapping sentences to semantic representations that resolve language-internal ambiguities. With this approach we hope to take a step forward in one of the main goals in artificial intelligence, namely the task of {\em natural language understanding}. In this paper, however, we emphasise the use of such models in the discovery of linguistic properties and relationships between languages in particular. Having that in mind, the study may open new directions for collaborations between language technology and general linguistics. But before coming back to this, let us first look at related work and the general principles of distributional semantics with cross-lingual grounding.

The use of translations for disambiguation has been explored in various studies. Dyvik \cite{Dyv02} proposes to use word translations to discover lexical semantic fields, Carpuat et al. \cite{carpuat:2013:SemEval-2013} discuss the use of parallel corpora for word sense disambiguation, van der Plas and Tiedemann \cite{vdPlasTiedemann:ACL06} present work on the extraction of synonyms and Villada and Tiedemann \cite{VilladaTiedemann:EACL06} explore multilingual word alignments to identify idiomatic expressions.

The idea of cross-lingual disambiguation is simple.
The following example illustrates the effect of disambiguation of idiomatic uses of {\em ``put off''} through translation into German:

\bigskip

\noindent{\footnotesize
\begin{tabular}{rlr}
English: & I don't want to {\bf put} you {\bf off}.   & The meeting has been {\bf put off} again.\\
German: & Ich will dich nicht {\bf abschrecken}.  & Das Treffen wurde wieder {\bf verschoben}.\\
Gloss: & {\em I will you not {\bf scare (off)}.}        & {\em The meeting has\_been again {\bf postponed}.}\\
\end{tabular}
}

\bigskip

Using the general idea of the distributional hypothesis that ``you shall know a word by the company it keeps'' \cite{Fir57}, we can now explore how cross-lingual context can serve as the source of information that defines the semantics of given sentences. As common in the field of distributional semantics, we will apply 
{\em semantic vector space models} that describe the meaning of a word or text by mapping it onto a position (a real-valued vector) in some high-dimensional Euclidean space. Various models and algorithms have been proposed in the literature (see, e.g., \cite{Erk:2012,Clark:2015}) and applied to a number of practical tasks. Predictive models based on neural network classifiers and neural language models \cite{BengioEA:2003,MikolovEA:2013a} have superseded models that are purely based on co-occurrence counts (see \cite{baroni-dinu-kruszewski:2014:P14-1} for a comparison of common approaches). Semantic vector spaces show even interesting algebraic properties that reflect semantic compositionality, support vector-based reasoning and can be mapped across languages 
\cite{mikolov-yih-zweig:2013:NAACL-HLT,MikolovEA:2013b}.
 Multilingual models have been proposed as well \cite{zou-EtAl:2013:EMNLP,klementiev-titov-bhattarai:2012:PAPERS}. Neural language models are capable of integrating multiple languages 
\cite{OstlingTiedemann:2017},
which makes it possible to discover relations between them based on the language space learned purely from the data.

\begin{figure}[t]
\includegraphics[height=5cm]{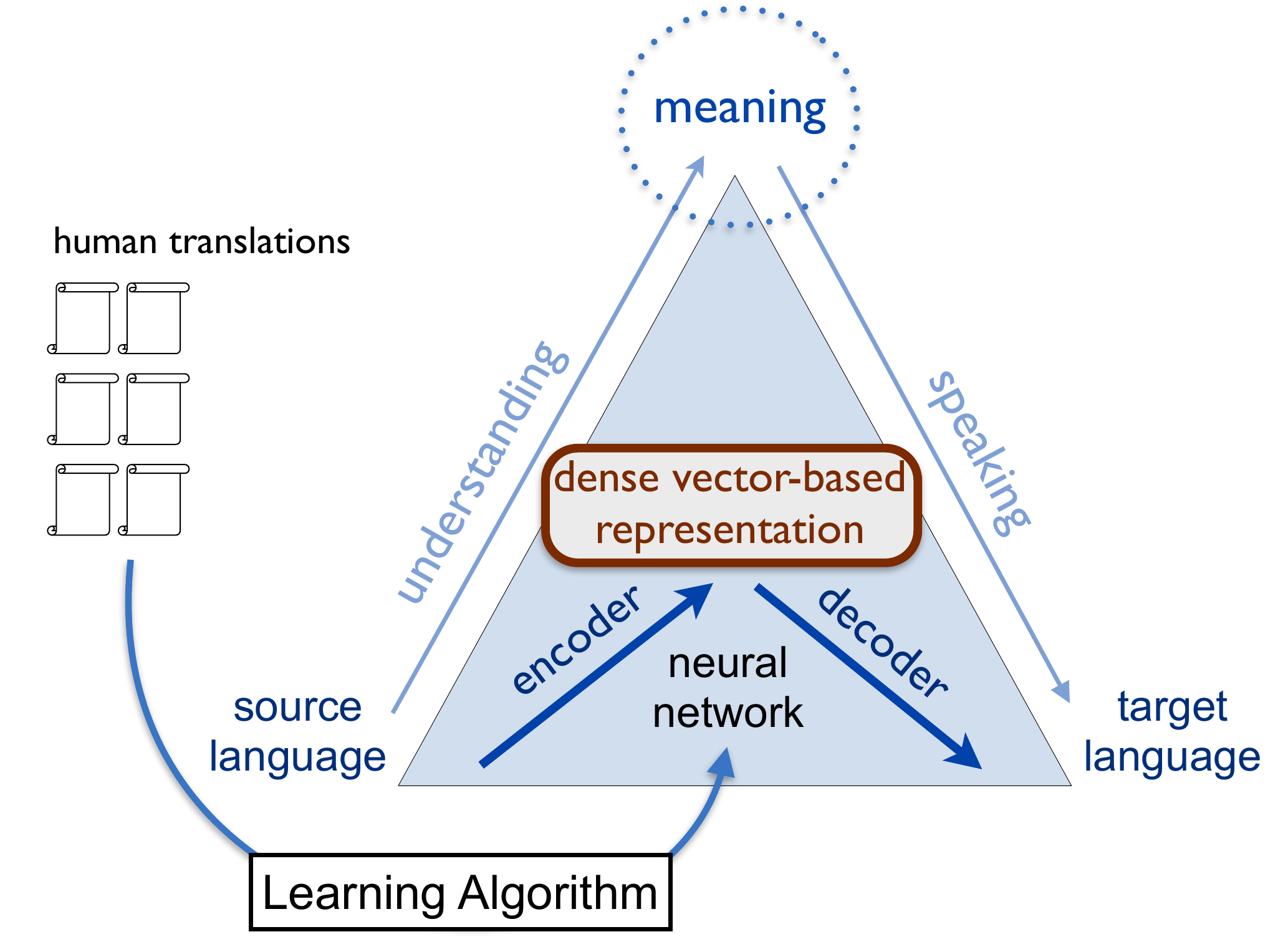}
\includegraphics[height=5cm]{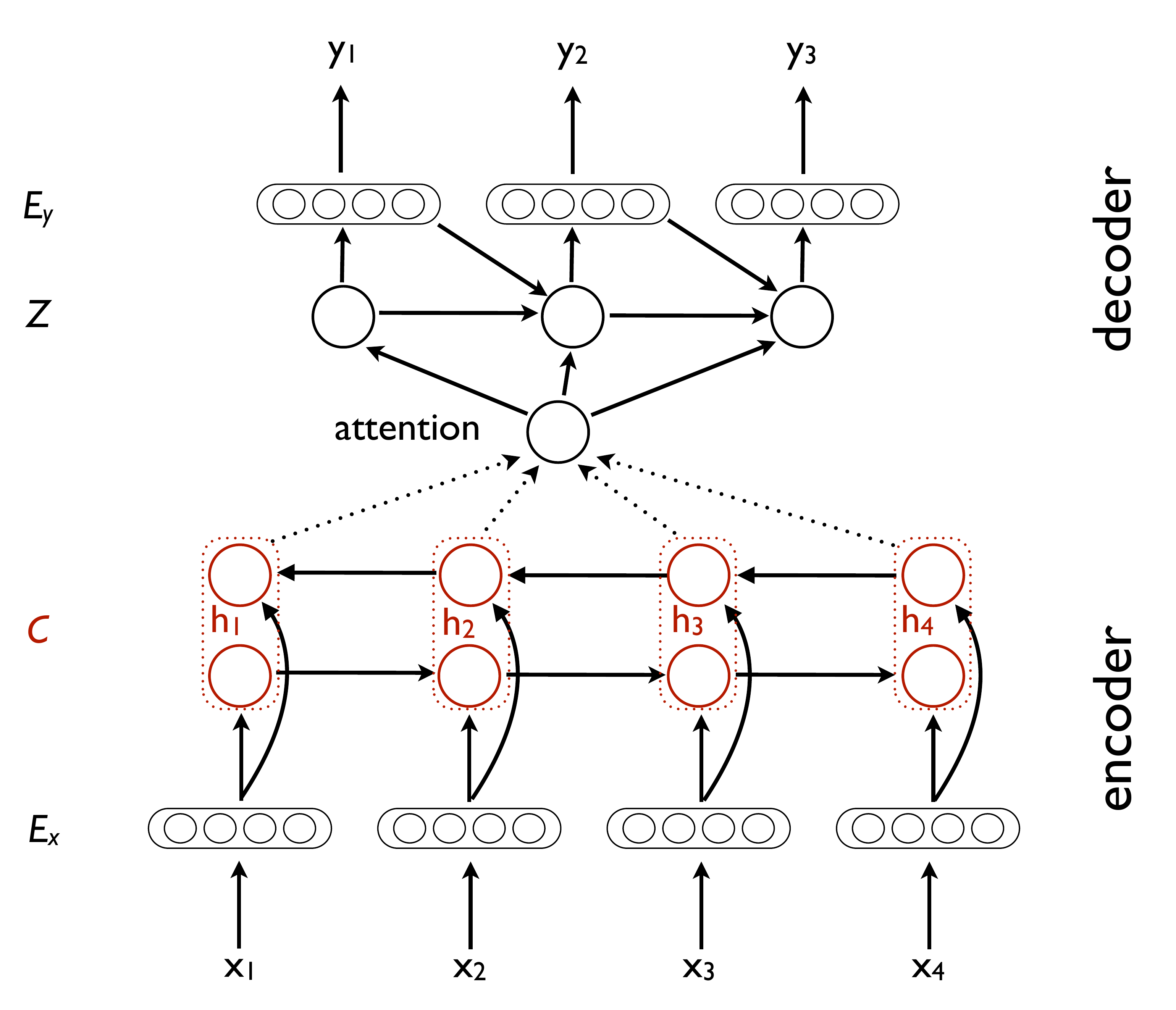}
\caption{Conceptual illustrations of neural machine translation and abstractions to meaning representations.}
\label{fig:neuralMT}
\end{figure}

Our framework will be neural machine translation (NMT) that applies an encoder-decoder architecture, which runs sequentially through a string of input symbols (for example words in a sentence) to map the information to dense vector representations, which will then be used to decode that information in another language. Figure~\ref{fig:neuralMT} illustrates the general principle with respect to the classical Vauquois triangle of machine translation \cite{Vauquois:76}.

Translation models are precisely the kind of machinery that tries to transfer the meaning expressed in one language into another by analysing (understanding) the input and generating the output. NMT tries to learn that mapping from data and, thus, learns to ``understand'' some source language in order to produce proper translations in a target language from given examples.
Our primary hypothesis is that we can increase the level of abstraction by including a larger diversity in the training data that pushes the model to improve compression of the growing variation and complexity of the task.
We will test this hypothesis by training multilingual models over hundreds or even almost a thousand languages to force the MT model to abstract over a large proportion of the World's linguistic diversity.

As a biproduct of multilingual models with shared parameters, we will obtain a mapping of languages to a continuous vector space depicting relations between individual languages by means of geometric distances. In this paper, we present our initial findings when training such a model with over 900 languages from a collection of Bible translations and focus on the ability of the model to pick up genetic relations between languages when being forced to cover many languages in one single model.

In the following, we will first present the basic architecture of the neural translation model together with the setup for training multilingual models. After that we will discuss our experimental results before concluding the paper with some final comments and prospects for future work.

\section{Multilingual Neural Machine Translation}

Neural machine translation typically applies an end-to-end network architecture that includes one or several layers for encoding an input sentence into an internal dense real-valued vector representation and another layer for decoding that representation into the output of the target language.
Various variants of that model have been proposed in the recent literature \cite{cho-EtAl:2014:SSST-8,bahdanau:ICLR:2015} with the same general idea of compressing a sentence into a representation that captures all necessary aspects of the input to enable proper translation in the decoder. 
An important requirement is that the model needs to support variable lengths of input and output. This is achieved using recurrent neural networks (RNNs) that naturally support sequences of arbitrary lengths. A common architecture is illustrated in Figure~\ref{fig:neuralMT}: 

Discrete input symbols are mapped via numeric word representations (embeddings $E$)
onto a hidden layer ($C$) of context vectors ($h$), in this case by a bidirectional RNN that reads the sequence in a forward and a reverse mode. The encoding function is often modeled by special memory units and all model parameters are learned during training on example translations. In the simplest case, the final representation (returned after running through the encoding layer) is sent to the decoder, which unrolls the information captured by that internal representation.
Note that the illustration in Figure~\ref{fig:neuralMT} includes an important addition to the model, a so-called attention mechanism. Attention makes it possible to focus on particular regions from the encoded sentence when decoding \cite{bahdanau:ICLR:2015} and, with this, the representation becomes much more flexible and dynamic and greatly improves the translation of sentences with variable lengths.

\begin{figure}[ht]
  \begin{minipage}[b]{0.51\textwidth}
    \includegraphics[height=5.3cm]{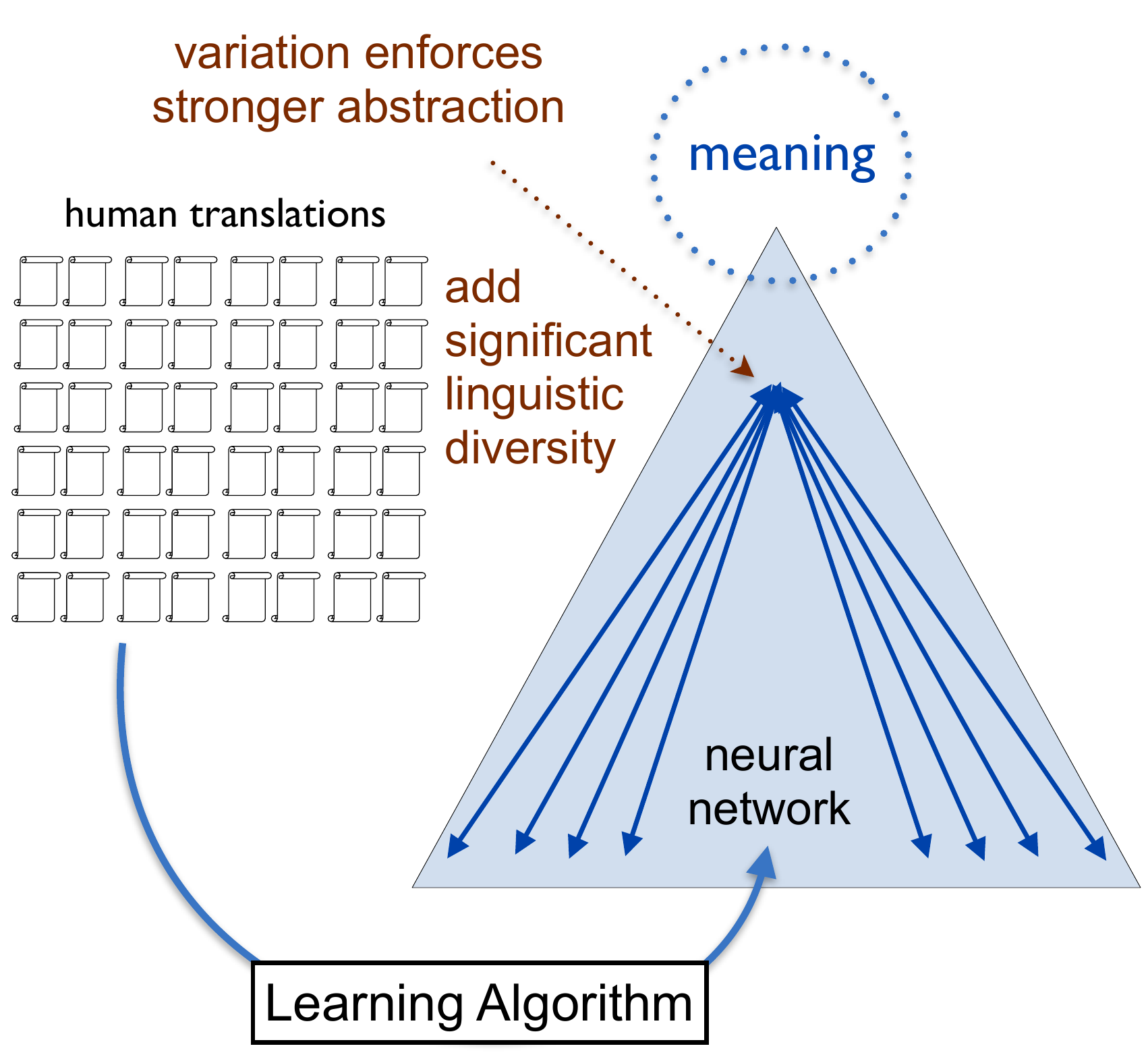}
  \end{minipage}
  \begin{minipage}[t]{0.49\textwidth}
    \noindent{\footnotesize\begin{tabular}[b]{ll}
              {\em Source:}	& DAN I don't know what to say. \\
              {\em Target:}       & Jeg ved ikke, hvad jeg skal sige.\\
& \\
              {\em Source:}	& SWE I don't know what to say. \\
              {\em Target:}        & Jag vet inte vad jag ska säga.\\
& \\
              {\em Source:}	& GER I don't know what to say. \\
              {\em Target:}	 &  Ich finde keine Worte. \\
& \\
              {\em Source:}	& DUT I don't know what to say. \\
              {\em Target:}	 &  Ik heb er geen woorden voor.\\
& \\
& \\
            \end{tabular}}
   \end{minipage}
\caption{Multilingual Neural MT and training data with language flags.}
\label{fig:multiNMT}
\end{figure}

All parameters of the network are trained on large collections of human translations (parallel corpora) typically by some form of gradient descent (iterative function optimisation) that is backpropagated through the network.
The attractive property of such a model is the ability to learn representations that reflect semantic properties of the input language through the task of translation. However, one problem is that translation models can be ``lazy'' and avoid abstractions if the mapping between source and target language does not require any deep understanding. This is where the idea of multilinguality comes into the picture: If the learning algorithm is confronted with a large linguistic variety then it has to generalize and to forget about language-pair-specific shortcuts. Covering substantial amounts of the world's linguistic diversity as we propose pushes the limits of the approach and strong abstractions in $C$ can be expected. Figure~\ref{fig:multiNMT} illustrates the intuition behind that idea.

Various multilingual extensions of NMT have already been proposed in the literature. The authors of \cite{DBLP:journals/corr/LuongLSVK15,dong-EtAl:2015:ACL-IJCNLP2} apply multitask learning to train models for multiple languages. Zoph and Knight \cite{DBLP:journals/corr/ZophK16} propose a multi-source model and \cite{DBLP:journals/corr/LeeCH16} introduces a character-level encoder that is shared across several source languages. 
In our setup, we will follow the main idea proposed by Johnson et al. \cite{DBLP:journals/corr/JohnsonSLKWCTVW16}. The authors of that paper suggest a simple addition by means of a language flag on the source language side (see Figure~\ref{fig:multiNMT}) to indicate the target language that needs to be produced by the decoder. This flag will be mapped on a dense vector representation and can be used to trigger the generation of the selected language. The authors of the paper argue that the model enables transfer learning and supports the translation between languages that are not explicitly available in training. 
This ability gives a hint of some kind of vector-based ``interlingua'', which is precisely what we are looking for. 
However, the original paper only looks at a small number of languages and we will scale it up to a larger variation using significantly more languages to train on. More details will be given in the following section.

\section{Experiments and Results}

\begin{figure}[ht]
\centering
\includegraphics[height=6cm ]{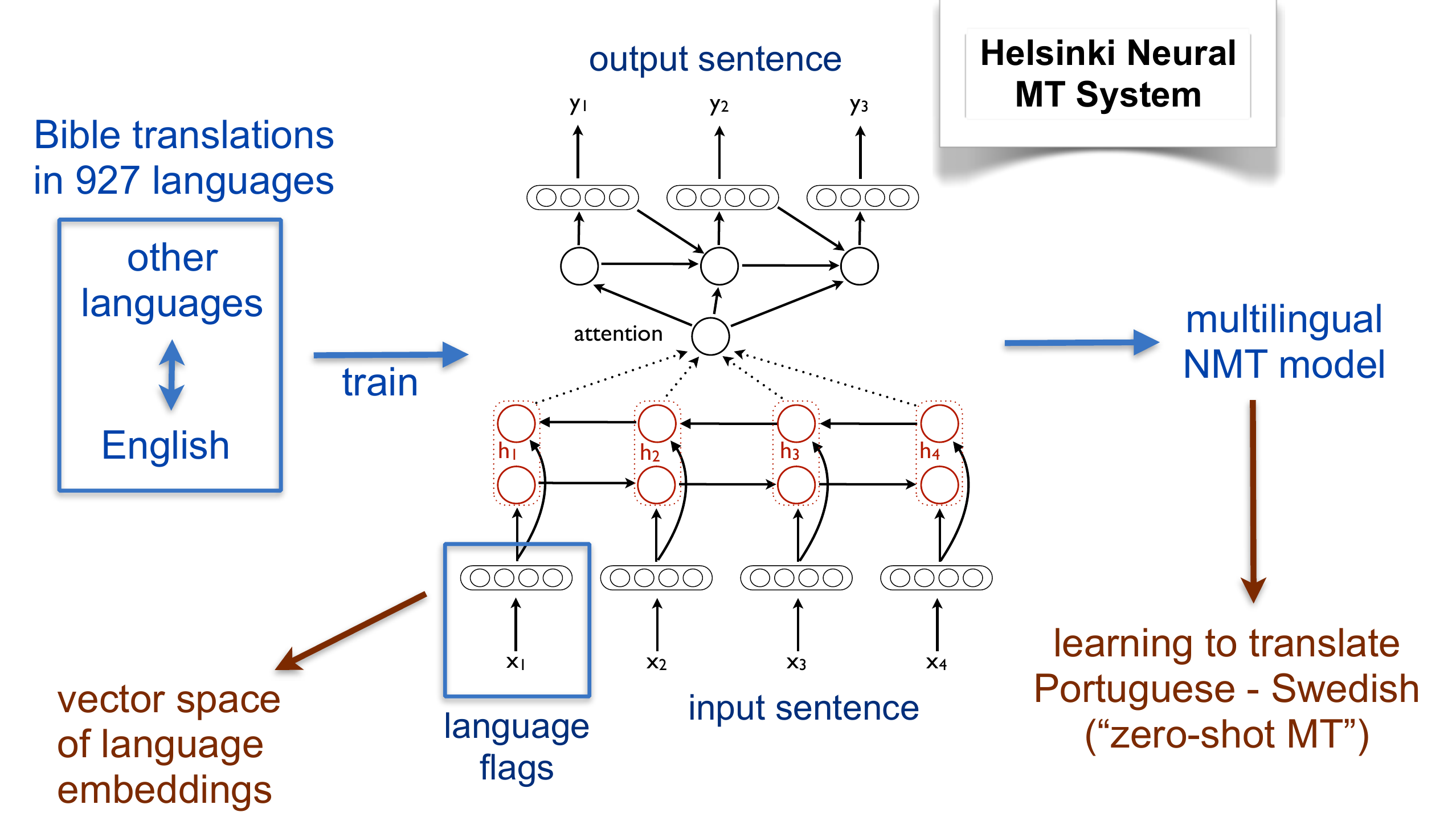}
\caption{Experimental setup: Bible translations paired with English as either source or target language are used to train one single multilingual NMT model including language flags on the source language side. Language flags are mapped onto a continuous vector space of {\em language embeddings}.}
\label{fig:setup}
\end{figure}

Our question is whether we can use a standard NMT model with a much larger coverage of the linguistic diversity of the World in order to maximise the variation signalling semantic distinctions that can be picked up by the learning procedures. 
Figure~\ref{fig:setup} illustrates our setup based on a model trained on over 900 languages from the multilingual Bible corpus \cite{MAYER14.220.L14-1215}.

We trained the model in various batches and observed the development of the model in terms of translation quality on some small heldout data. The heldout data refers to an unseen language pair, Swedish-Portuguese in our case (in both directions). We selected those languages in order to see the capabilities of the system to translate between rather distant languages for which a reasonable number of closely related languages are in the data collection to improve knowledge transfer.

\begin{figure}[ht]
\centering
\includegraphics[width=.9\textwidth]{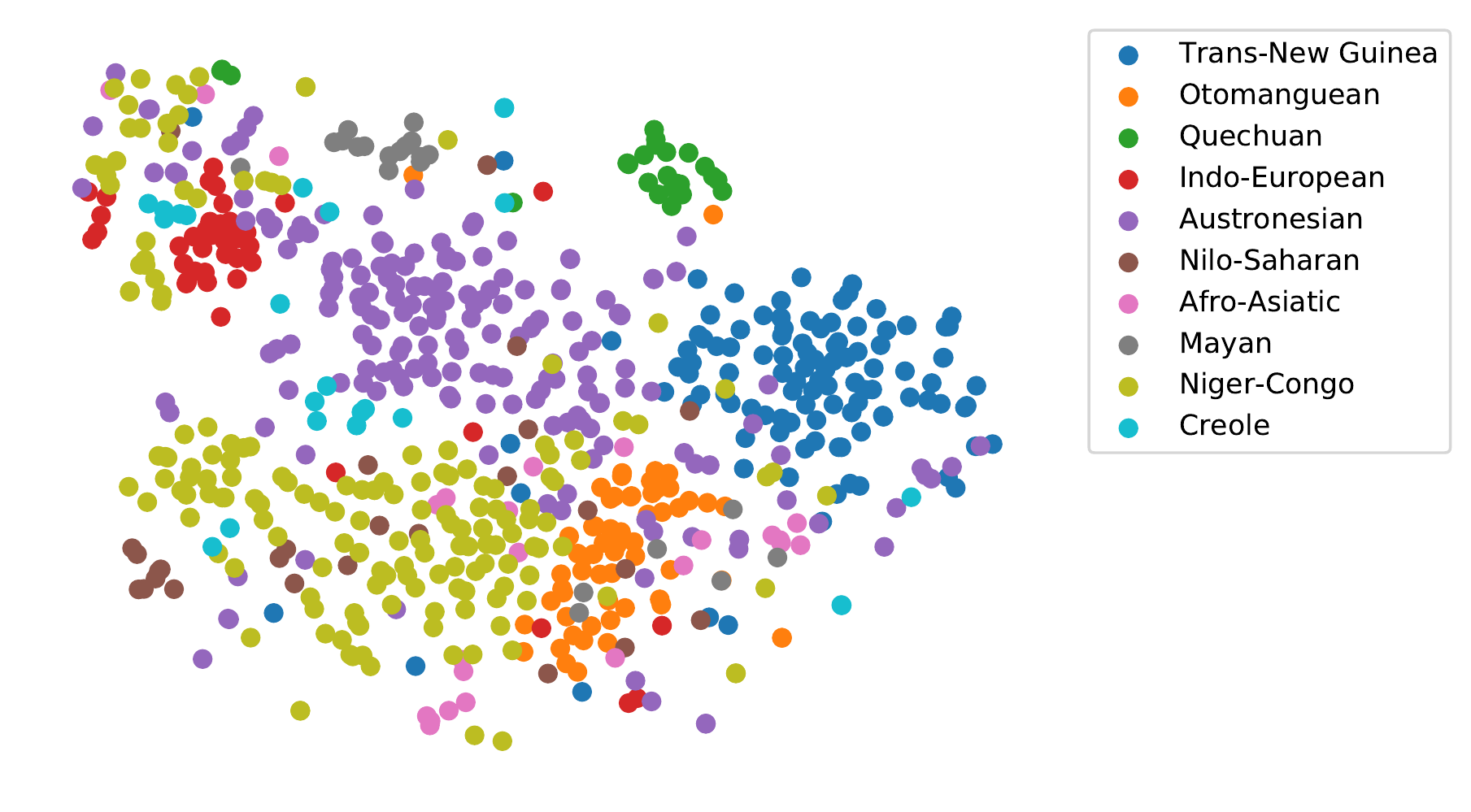}
\caption{Continuous language space that emerges from multilingual NMT (t-SNE plot).}
\label{fig:langspace}
\end{figure}

The results demonstrate so far that the network indeed picks up the information about the language to be produced. The decoder successfully switches to the selected language and produces relatively fluent Bible-style text. The adequacy of the translation, however, is rather limited and this is most probably due to the restricted capacity of the network with such a load of information to be covered. Nevertheless, it is exciting to see that such a diverse material can be used in one single model and that it learns to share parameters across all languages. One of the most interesting effects that we can observe is the emerging language space that relates to the language flags in the data. In Figure~\ref{fig:langspace} we plot the language space (using t-SNE \cite{MaatenHinton:2008} for projecting to two dimensions) coloured by language family for the ten language families / groups with most members in our data set. 

We can see that languages roughly cluster according to the family they belong to. Note that this is purely learned from the data based on the objective to translate between all of them with a single model. The training procedure learns to map closely related languages near to each other in order to increase knowledge transfer between them. This development is very encouraging and demonstrates the ability of the neural network model to optimise parameter sharing to make most out of the model's capacity.

An interesting question coming out of this study is whether such multilingual translation models can be used to learn linguistic properties of the languages involved. Making it possible to measure the distance between individual languages in the emerging structures could be useful in data-driven language typology and other cross-linguistic studies. The results so far, do not reveal a lot of linguistically interesting relations besides the projection of languages onto a global continuous space with real-values distances between them. Nevertheless, quantifying the distance is potentially valuable and provides a more fine-grained relation than discrete relations coming from traditional family trees. It is, however, still an open question what kind of properties are represented by the language embeddings and further studies are necessary to see whether specific linguistic features can be identified and isolated from the distributed representations. There is a growing interest in interpretability of emerging structures and related work already demonstrates the ability of predicting typological features with similar language representations \cite{bjerva-augenstein:2018:IWCLUL}.

Massively parallel data sets make it now possible to study specific typological structures with computational models, for example tense and aspect as in \cite{asgari-schutze:2017:EMNLP2017}, and we intend to follow up our initial investigations of NMT-based representations in future research along those lines. We also plan to consider other domains than the one of religious texts but it is difficult to obtain the same coverage of the linguistic space with different material. Unbalanced mixtures will be an option but difficult to train. Resources like the Universal Declarations of Human Rights are an option but, unfortunately, very sparse.

Another direction is to explore the inter-lingual variations and language developments using, for example, the alternative translations that exist for some languages in the Bible corpus. However, even here the data is rather sparse and it remains to be seen how reliable any emerging pattern will be. Crucial for the success will be a strong collaboration with scholars from the humanities, which shows the important role of digital humanities as a field.


\section{Conclusions}

In this paper, we present our experiments with highly multilingual translation models. We trained neural MT models on Bible translations of over 900 languages in order to see whether the system is capable of sharing parameters across a large diverse sample of the World's languages. Our motivation is to learn language-independent meaning representations using translations as implicit semantic supervision and cross-lingual grounding. Our pilot study demonstrates that such a model can pick up the relationship between languages purely from the data and the translation objective. We hypothesise that such a data-driven setup can be interesting for cross-linguistic studies and language typology. In the future, we would like to investigate the emerging language space in more detail also in connection with alternative network architectures and training procedures. We believe that empirical methods like this one based on automatic representation learning will have significant impact on studies in linguistics providing an objective way of investigating properties and structures of human languages emerging from data and distributional patterns.

\section*{Acknowledgements}

We would like to thank the anonymous reviewers for their valuable comments and suggestions as well as the Academy of Finland for the support of the research presented in the paper with project 314062 from the ICT 2023 call on Computation, Machine Learning and Artificial Intelligence.


\bibliographystyle{splncs}
\bibliography{references}

\end{document}